\documentclass{article}

\usepackage{arxiv}

\usepackage[utf8]{inputenc} 
\usepackage[T1]{fontenc}    
\usepackage{hyperref}       
\usepackage{url}            
\usepackage{booktabs}       
\usepackage{amsfonts}       
\usepackage{nicefrac}       
\usepackage{microtype}      
\usepackage{lipsum}
\usepackage{graphicx}
\usepackage{amssymb, amsmath}
\usepackage{color}
\usepackage{url}
\usepackage{comment}
\usepackage{afterpage}
\usepackage{here}
\usepackage{mathrsfs}

\newcommand{\reffig}[1]{Figure~\ref{#1}}

\newcommand{\reftab}[1]{Table~\ref{#1}}
\newcommand{\refeq}[1]{\eqref{#1}}

\newcommand{\bi}[1]{\ensuremath{\boldsymbol{#1}}}

\makeatletter
\newcommand{\figcaption}[1]{\def\@captype{figure}\caption{#1}}
\newcommand{\tblcaption}[1]{\def\@captype{table}\caption{#1}}
\makeatother

\title{Physics-guided generative adversarial network to learn physical models}

\author{
	Kazuo Yonekura \thanks{\texttt{yonekura@struct.t.u-tokyo.ac.jp}}\\
	Department of Systems Innovasion\\
	The University of Tokyo\\
	Tokyo, JAPAN 113-8656 \\
}

\begin{document}
	\maketitle
	
	\begin{abstract}
This short note describes the concept of guided training of deep neural networks (DNNs) to learn physically reasonable solutions. DNNs are being widely used to predict phenomena in physics and mechanics. One of the issues of DNNs is that their output does not always satisfy physical equations. One approach to consider physical equations is adding a residual of equations into the loss function; this is called physics-informed neural network (PINN). One feature of PINNs is that the physical equations and corresponding residual must be implemented as part of a neural network model. In addition, the residual does not always converge to a small value. The proposed model is a physics-guided generative adversarial network (PG-GAN) that uses a GAN architecture in which physical equations are used to judge whether the neural network's output is consistent with physics. The proposed method was applied to a simple problem to assess its potential usability.
	\end{abstract}


\section{Introduction}
Deep neural networks are widely employed in diverse research areas including physics and engineering. 
In mechanical engineering, DNNs are often used as a surrogate model \cite{Sun20,Tripathy18}. 
In general, a DNN learns from data, and the output is not guaranteed to be consistent with physics even if the data were generated by certain physical models. 

Generative adversarial networks (GANs) \cite{Goodfellow14}, constitute a type of DNN used to solve inverse problems \cite{Achour20,Yonekura21a,Yonekura22c}.  
In these previous studies, the GAN models were trained using a set of data consisting of airfoil shapes and their lift coefficient at a certain angle of attack. Then, by inputting a specific lift coefficient, the trained model output airfoil shapes associated to that input lift coefficient. 
The lift coefficients of the generated shapes were close to the specified label, but some errors were also identified. 
These errors were due to the fact that the lift coefficient was calculated from the airfoil shapes, but the physical equations were not input to the DNN model. 

To consider physical consistency, physics-informed neural networks (PINNs) have been proposed. One approach for PINNs is to add the residual of physical equations to the loss function of the DNN \cite{Sun20,willard22}. 
PINNs were used to predict various targets such as lake water level \cite{Jia21}, surface water level \cite{Bertels23}, and seismic response \cite{Zhang20}. 
However, a PINN model needs the physical equations to be implemented in a DNN architecture. 
This causes difficulties from an application point of view. 
For example, commercial software cannot be used in PINN models. 
Hence, it is desirable to consider arbitrary physical equations and commercial software in DNN surrogate models and generative models. 

The proposed model aims to consider arbitrary physical equations using a GAN architecture. 
GANs were proposed by \cite{Goodfellow14}. This architecture consists of a generator network and a discriminator network. 
The generator network outputs data that mimic training data, whereas the discriminator network distinguishes generated data from training data. 
Training data are referred to as true data, whereas generated data constitute fake data. 
In the proposed PG-GAN model, true or fake is defined by physical equations; if the residual is smaller than a specific value, then the generated data are true; otherwise, the generated data are fake. 
The physical model guides the DNN to learn physical consistency and is only used to categorize data as true or fake. 
The physical equations remain outside of the DNN model and are not implemented in it. 
Therefore, arbitrary physical models can be used. 
By decreasing the threshold value, the residual of the generated data is decreased. 
In the PINN model, the residual is added to the loss function and cannot be controlled. 

The present paper is organized as follows. 
GAN and PINN models are explained in Section 2. 
A PG-GAN is formulated in Section 3. 
A numerical study conducted on a simple problem is presented in Section 4.
Conclusions are provided in Section 5.

\section{GAN and Physics-Informed GAN}
\begin{figure*}[htb]
	\begin{center}
		\begin{minipage}[h]{\textwidth}
			\begin{center}
				\scalebox{0.15}{\includegraphics[]{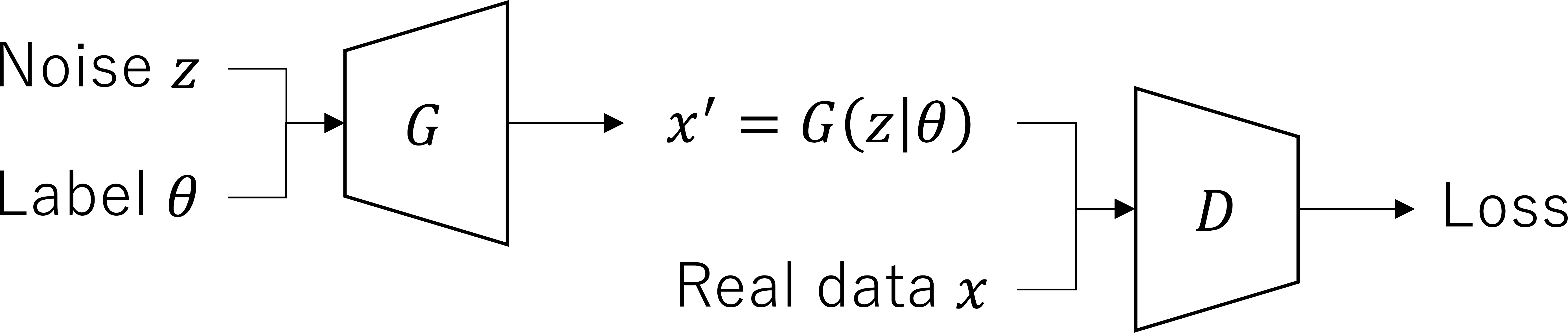}}
			\end{center}
		\end{minipage}%
		\caption{Architecture of GAN.}
		\label{fig:GAN}
	\end{center}
\end{figure*}

A conditional GAN model consists of a generator network $G$ and a discriminator network $D$, as illustrated in \reffig{fig:GAN}. 
The input of the generator network is a noise vector $\bi{z}$ and label $\theta$, and the output represents fake data $\bi{x}'$ expressed as $\bi{x}' = G(\bi{z} \mid \theta) $.
The input of the discriminator network is given by real $\bi{x}$ and fake $\bi{x}'$ data, and the network distinguishes real data from fake data. 
The loss function is defined as
\begin{align}
	V(G,D) 
	= 
	E_{\bi{x} \sim p_x} \left[ \log D(x) \right] 
	+ E_{\bi{z} \sim p_z} \left[ \log \left( 1 - D \left( G(z) \right) \right) \right] , \label{eq.V}
\end{align}
and the generator minimizes $V$ whereas the discriminator maximizes $V$; 
i.e., $	\min_G \max_D V(G,D). $
The discriminator only considers data $\bi{x}$ and $\bi{x}'$; physical reasonableness is not considered.

A PINN can be coupled with a GAN model. The resulting architecture is called PI-GAN in the present article. 
Suppose that a physical model is expressed as $P(\bi{x}) = 0$. For a variable $\bi{x}'$, the residual is given by $r=P(\bi{x}')$. 
The PINN adds the residual into the loss function which is minimized. In the GAN model, the loss function of the generator is modified as 
$V(G,D) + \lambda r$, where $\lambda$ is a constant. In the numerical example described later on, $\lambda$ was set to $0.1$. 
The loss function of the discriminator remains the same as the original GAN.

\section{Proposed PG-GAN model}
Generated data should be considered as true if they are physically reasonable. 
To judge physical reasonableness, a physical model is used. 
Suppose that the physical model is described as 
$
P(\bi{x} \mid \theta) =0, 
$
and the value of $P( \bi{x} \mid \theta)$ is treated as a residual. 
We treat data $\hat{\bi{x}}$  as true if $P( \hat{\bi{x}} \mid \hat{\theta} ) \leq \varepsilon$. 
Hence, the input of our discriminator is given by generated data $\bi{x}'$ and the output is whether data are physically reasonable or not. 
For a given $\varepsilon$, let a set $ \mathcal{D}_{\varepsilon}$ represent a set of data whose residual is equal to, or less than, $\varepsilon$. 
\begin{align*}
	\mathcal{R}_{\varepsilon} = \left \{  \bi{x} \mid P \left( \bi{x} \mid \theta \right) \leq \varepsilon \right\}
	,~
	\mathcal{F}_{\varepsilon} = \left \{  \bi{x} \mid P \left( \bi{x} \mid \theta \right) > \varepsilon \right\}
\end{align*}
In this case, the loss function becomes
\begin{align}
	V_\varepsilon (G,D) 
	= E_{  G(\bi{z}) \in \mathcal{R}_\varepsilon  } \left[ \log D(x) \right] 
	+ E_{ G(\bi{z}) \in \mathcal{F}_\varepsilon } \left[ \log \left( 1 - D \left( G(z) \right) \right) \right] . \label{eq.Ve}
\end{align}
The optimization problem for the discriminator is 
$
\max_{D} V_\varepsilon (G,D).
$
The discriminator tries to mimic the physical model to judge physical reasonableness.

If we minimize $V_\varepsilon (G,D)$ with respect to $G$, the generator is not trained as desired. 
In the ordinal GAN loss function, the first term of $V(G,D)$  is not a function of $G$ and the generator optimization problem becomes $ \min_{G} ~ E_{\bi{z} \sim p_z} \left[ \log \left( 1 - D \left( G(z) \right) \right) \right]$. 
However, in \refeq{eq.Ve}, both the first and second terms were functions of $G$. 
Therefore, the generator optimization problem uses only the second term of $V_\varepsilon (G,D)$ and is 
$\min_{G}  E_{ G(\bi{z}) \in \mathcal{F}_\varepsilon } \left[ \log \left( 1 - D \left( G(z) \right) \right) \right], $ 
instead of $ \min_G ~ V_\varepsilon (G,D)$. 

\begin{figure*}[h]
	\begin{center}
		\begin{minipage}[h]{\textwidth}
			\begin{center}
				\scalebox{0.15}{\includegraphics{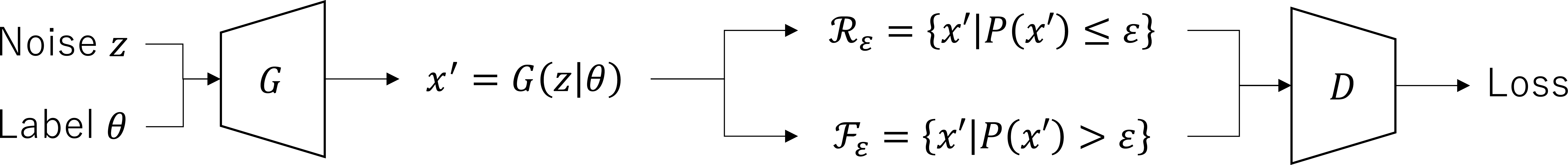}}
			\end{center}
		\end{minipage}%
		\caption{Architecture of PG-GAN.}
		\label{fig:PGGAN}
	\end{center}
\end{figure*}

The architecture of the proposed model is illustrated in \reffig{fig:PGGAN}. 
The physics-guided GAN model uses a physics model as a referee to judge whether the generated data are physically reasonable or not.  
The discriminator is a surrogate model of the physical model. 
If we could use the physical model itself as a discriminator, the generator would be trained much more efficiently. 
However, if we use the physical model in the architecture, back propagation would stop at the physical model because we are assuming that the physical model is a black-box software. 
This is also a feature of the PG-GAN: the training data no longer appear in the model. The real data are not necessary because true/false is judged by the physical model.

The PINN can be coupled with the PG-GAN by adding the residual into the loss function of the generator; the resulting architecture is called PG-PI-GAN. The loss function is modified as 
$
\min_{G}  E_{ G(\bi{z}) \in \mathcal{F}_\varepsilon } \left[ \log \left( 1 - D \left( G(z) \right) \right) \right] + \lambda r. 
$

Training the PG-GAN without pre-training is not efficient because in the early epochs, the generator cannot generate physically reasonable data and $ \mathcal{R}_\varepsilon$ always becomes an empty set. 
In such a case, both the generator and discriminator are not trained well because the discriminator always outputs $0$ (fake) whereas the generator has no clue to generate reasonable data. 
Hence, it is necessary to start from a pre-trained generator that generates non-empty $ \mathcal{R}_\varepsilon$.  
To obtain such a pre-trained generator, an ordinal GAN model is used. 

Data generated by the pre-trained generator exhibits a large residual of physical equation $P(\bi{x}' \mid \theta)$. 
$\varepsilon$ must be large enough so that both sets $\mathcal{R}_\varepsilon$ and $\mathcal{F}_\varepsilon$ are not empty sets. 
However, it is not desirable to terminate with large $\varepsilon$. 
Hence, $\varepsilon$ is reduced as training proceeds until it reaches the target value $\bar{\varepsilon}$. 
In the following numerical example, $\varepsilon$ was constant for 10,000 epochs and then changed. 
Alternatively, $\varepsilon$ can be gradually reduced. 

\section{Numerical study}
\begin{figure}[h]
	\begin{minipage}[c]{\textwidth}
	\centering
	\scalebox{0.6}{\includegraphics{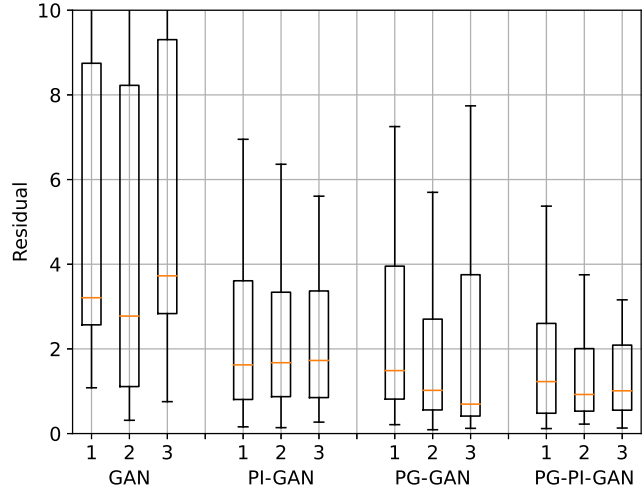}}
	\caption{Convergence of PG-GAN.}
	\label{fig:convergence}
\end{minipage}
\end{figure}

Newton's equation of motion under gravity is expressed as 
$
\bi{x}(t) = \bi{x}_0 -\frac{1}{2} \bi{g} t^2 + \bi{v}_0 t, 
$
where $\bi{x}_0$, $\bi{g}$, $\bi{v}_0$, and $t$ denote the coordinate of an initial point, gravitational acceleration, initial velocity vector, and time, respectively. 
Physical equation $P$ is formulated as 
$ P(\bi{x}(t), \theta=\bi{v}_0) = \left \| \bi{x}(t) - \bi{x}_0 + \frac{1}{2} \bi{g} t^2 - \bi{v}_0 t \right \|^2, $ 
where $\bi{v}_0 = v_0 \left(  \cos{\phi}, \sin{\phi} \right)^{\top}$.
The task is to output a sequence of coordinates 
$ \bi{\xi}(\theta=\bi{v}_0) = \left( \bi{x}(\delta_t; \theta)^{\top}, \bi{x}(2\delta_t; \theta)^{\top}, \dots, \bi{x}(100 \delta_t; \theta )^{\top} \right)^{\top}, $ 
where the parameter $\bi{v}_0$ is given. 
A dataset for pre-training is first prepared as
$ \mathcal{D}= \left \{ \bi{\xi}(\theta = \bi{v}_0)  \mid v_0 \in \{1, 2, \dots 100 \},~ \phi \in \{ 0, 1, \dots, 90 \}  \right \}.$  
The total number of training data was 9,000. 
In the pre-training, the GAN model was trained using dataset $\mathcal{D}$ for 10,000 epochs. 
Then, PG-GAN training was carried out. 
Threshold $\varepsilon$ is defined as a function of the number of epochs $e$ as follows:
\begin{align*}
	\varepsilon = 
	\left \{
	\begin{array}{llll}
		5, & {\rm (if~} 10,000 \leq e < 20,000 {\rm )},  \qquad &
		2.5, & {\rm (if~} 20,000 \leq e < 30,000 {\rm )}, \\
		1.25, & {\rm (if~} 30,000 \leq e < 70,000 {\rm )}, \qquad &
		0.625, & {\rm (if~} 70,000 \leq e < 100,000 {\rm )}, 
	\end{array} \right.
\end{align*}
The PG-GAN model was trained and data $\xi$ were output. 
The residual of the output coordinates was calculated by
$r = \frac{1}{100}\sum_{k=1}^{k=100} P(\bi{x}(k \delta_t), \theta=\bi{v}_0). $ 
Ordinal GAN, physics-informed GAN (PIGAN), and PG-PI-GAN were also trained and compared. 
Each model was trained and evaluated three times separately. 
\reffig{fig:convergence} shows the boxplot of residuals of each model. 
Data outside of the 1.5 interquartile range (IQR) from the first and third quartiles was treated as outliers in the boxplot.
\reftab{tab:stat} shows the average median, first quartile, and interquartile range for each model. 
The physics-informed GAN featured the same GAN architecture except for the residual $r$, which was added to the loss function. 
All network structures were the same in all models. 
The PG-GAN was characterized by lower median than the PI-GAN and by first quartile values.  
The PG-PI-GAN presented similar median and first quartile values as those of the PG-GAN, but the inter IQR was lower than that of the PG-GAN. 
These results show that the PG-GAN effectively reduces the median value, but does not reduce the IQR value. 
This difference comes from the loss functions of both models. 
The residual was added in the loss function of the PI-GAN, and hence the residual of all generated data was reduced. 
By contrast, the PG-GAN considered no residual in the loss function. The undesired data indicate larger residuals. Note that the amount of residuals of undesired data does not affect the loss function. Hence, the residuals of undesired data tend to become large. 
Therefore, the proposed PG-PI-GAN, which couples a PI-GAN and a PG-GAN, successfully reduces the median, first quartile, and IQR. 

\begin{table}
		\begin{center}
			\begin{tabular}{l|r|r|r|r}
				& GAN 		& PI-GAN & PG-GAN & PG-PI-GAN\\
				\hline
				Median    		& 3.24		& 1.67 	& 1.07 	& 1.05 \\
				First quartile 	& 2.17		& 0.84 	& 0.60 	& 0.52 \\
				Interquartile range
				& 6.59		& 2.60	& 2.88	&1.71 
			\end{tabular}
		\end{center}
		\caption{Statistics of residuals for Newton's equation of motion.}
		\label{tab:stat}
\end{table}

\section{Conclusions}
This short note describes a concept of PG-GAN.  
This PG-GAN uses arbitrary physical models, regardless of differentiability and smoothness, to guide neural networks to output physically reasonable solutions. 
One advantage of the proposed PG-GAN is that the physical model is outside of the neural network calculation graph; back-propagation in the neural network is not conducted on the physical model. Hence, any physical model can be utilized. For example, a commercial software could be used and one does not need to implement the physical model. 
Existing PINN models require physics equations to be implemented inside the calculation graph. Hence, arbitrary physics equations cannot be used; e.g., commercial software cannot be used. 
The proposed PG-GAN network does not need training data. The generator creates data and the physical model judges whether the output is reasonable or not. 
However, the PG-GAN model is pre-trained using an ordinal GAN model with training data. 

The proposed PG-GAN model was tested using Newton's equation of motion. The PG-GAN and PG-PI-GAN featured lower median values of residuals. 
When the PG-GAN was coupled with the PI-GAN, the IQR value also decreased.

\section*{Acknowledgements}
This study was supported by JSPS KAKENHI Grant Numbers JP21K14064 and JP23K13239.

\bibliographystyle{abbrv}  
\bibliography{bib-DDD}



\end{document}